\documentclass[letterpaper, 10pt, conference]{ieeeconf}  % Comment this line out if you need a4paper

\IEEEoverridecommandlockouts                              % This command is only needed if 
\usepackage{times}
\usepackage{epsfig}
\usepackage{color} 
\usepackage{amsmath}
\usepackage{amssymb}
\usepackage{graphicx}
\usepackage{multirow}
\usepackage{tabularx}
\usepackage{adjustbox}
\usepackage{siunitx}
\usepackage{textcomp}
\usepackage{array}
\usepackage{pifont}
\usepackage{booktabs}
\usepackage{amsfonts}
\usepackage[ruled]{algorithm2e}
\usepackage{hyperref}
\usepackage{cleveref}
\crefname{table}{Table}{Tables}
\Crefname{table}{Table}{Tables}

\usepackage{stmaryrd} % short arrow
\usepackage{latexsym} % arrow
\usepackage{url}
\usepackage{graphicx}
\usepackage{subcaption} % subcaption 패키지 추가
\usepackage{booktabs}
\usepackage{algpseudocode}

\newcommand{\figref}[1]{Fig.~\ref{#1}}

\newcommand{\tabref}[1]{Tab.~\ref{#1}}
\newcommand{\eqnref}[1]{Eq.~(\ref{#1})}

\newcommand{\secref}[1]{Sec.~\ref{#1}}

\newcommand{\ie}{\textit{i.e.}}
\newcommand{\eg}{\textit{e.g.}}
\newcommand{\etal}{\textit{et al.}}

\usepackage{xcolor,colortbl}
\definecolor{Gray1}{gray}{0.85}
\definecolor{Gray2}{gray}{0.65}

\usepackage{pifont}% http://ctan.org/pkg/pifont
\title{\LARGE \bf

Learning Quadrupedal Locomotion with Impaired Joints \\ Using Random Joint Masking 

}

\author{Mincheol Kim$^{1}$, Ukcheol Shin$^{2}$, Jung-Yup Kim$^{1}$% <-this % stops a space
\thanks{$^{1}$M. Kim and J. Kim are with the Humanoid Robot Research Laboratory, Department of Mechanical Design and Robot Engineering, Seoul National University of Science and Technology, Seoul, 01811, Republic of Korea
        {\tt\small \{kmc96, jyk76\}@seoultech.ac.kr}}%
\thanks{$^{2}$ U. Shin is with Robotics Institute, Carnegie Mellon University, Pittsburgh, PA, 15217, US
        {\tt\small ushin@andrew.cmu.edu}}%
}

\begin{document}

\maketitle
\thispagestyle{empty}
\pagestyle{empty}

%%%%%%%%%%%%%%%%%%%%%%%%%%%%%%%%%%%%%%%%%%%%%%%%%%%%%%%%%%%%%%%%%%%%%%%%%%%%%%%%
\begin{abstract}

Quadrupedal robots have played a crucial role in various environments, from structured environments to complex harsh terrains, thanks to their agile locomotion ability. 
However, these robots can easily lose their locomotion functionality if damaged by external accidents or internal malfunctions.
In this paper, we propose a novel deep reinforcement learning framework to enable a quadrupedal robot to walk with impaired joints.
The proposed framework consists of three components: 1) a random joint masking strategy for simulating impaired joint scenarios, 2) a joint state estimator to predict an implicit status of current joint condition based on past observation history, and 3) progressive curriculum learning to allow a single network to conduct both normal gait and various joint-impaired gaits.
We verify that our framework enables the Unitree's Go1 robot to walk under various impaired joint conditions in real-world indoor and outdoor environments.
\end{abstract}

%%%%%%%%%%%%%%%%%%%%%%%%%%%%%%%%%%%%%%%%%%%%%%%%%%%%%%%%%%%%%%%%%%%%%%%%%
\section{Introduction}
\label{sec:intro}

Recently, quadrupedal robots have played a crucial role in various applications, such as human rescue, disaster response, and harsh terrain exploration \cite{ref1,ref2,ref3}. 
To perform these applications in complex and dynamic environments, agile locomotion is a fundamental requirement for quadruped robots \cite{ref4}. 
However, the locomotion ability can be impaired by external accidents, such as hit-by-obstacle and impact-by-objects, and internal hardware issues, such as joint malfunction and locking.
These accidents and issues are more likely to occur in challenging and extreme environments.

\begin{figure}[t]
\centering % 중앙 정렬
\begin{tabular}{c}
\includegraphics[width=1.0\linewidth]{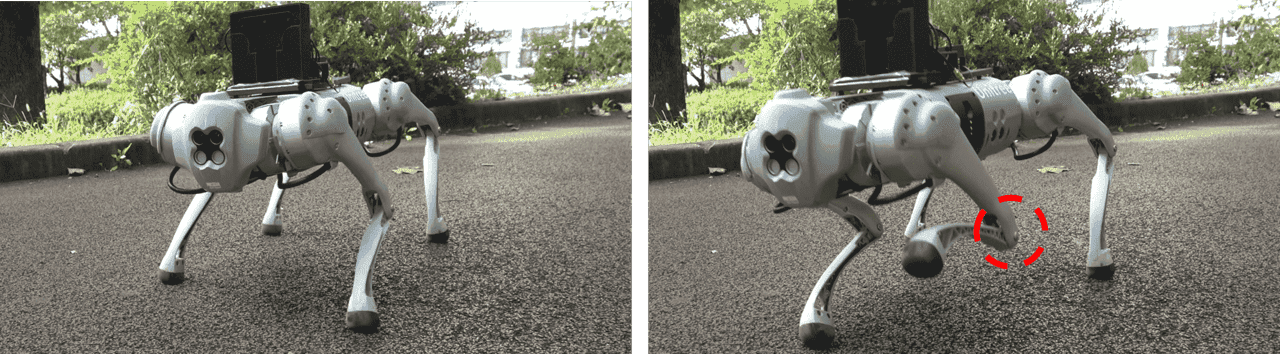} \vspace{-0.05in} \\ % 이미지 추가
{\footnotesize (a) Impaired joint conditions} \\
\includegraphics[width=1.0\linewidth]{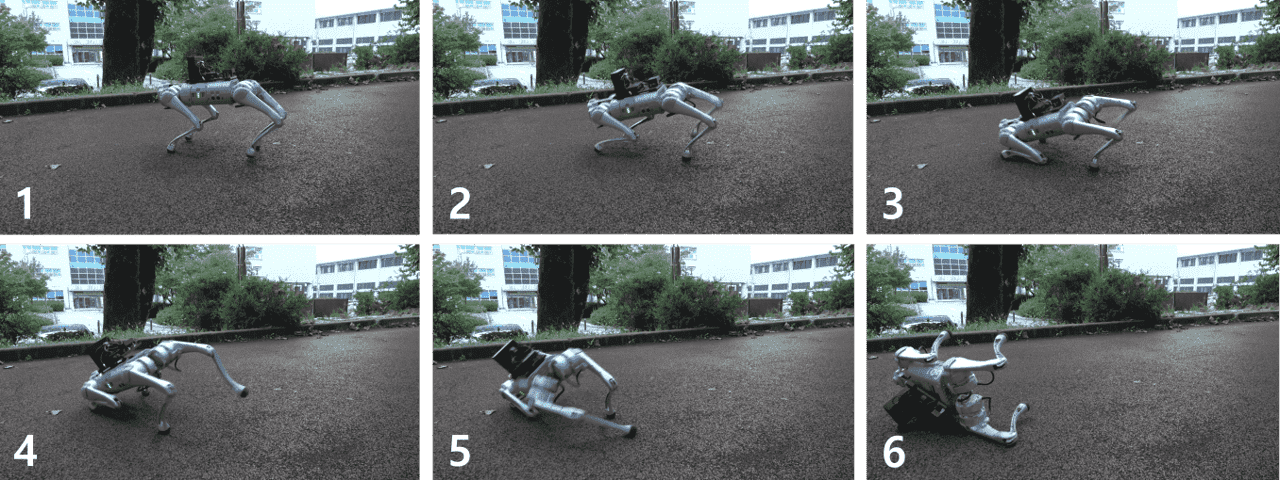} \vspace{-0.05in} \\  % 이미지 추가
{\footnotesize (b) Baseline~\cite{ref10}} \\
\includegraphics[width=1.0\linewidth]{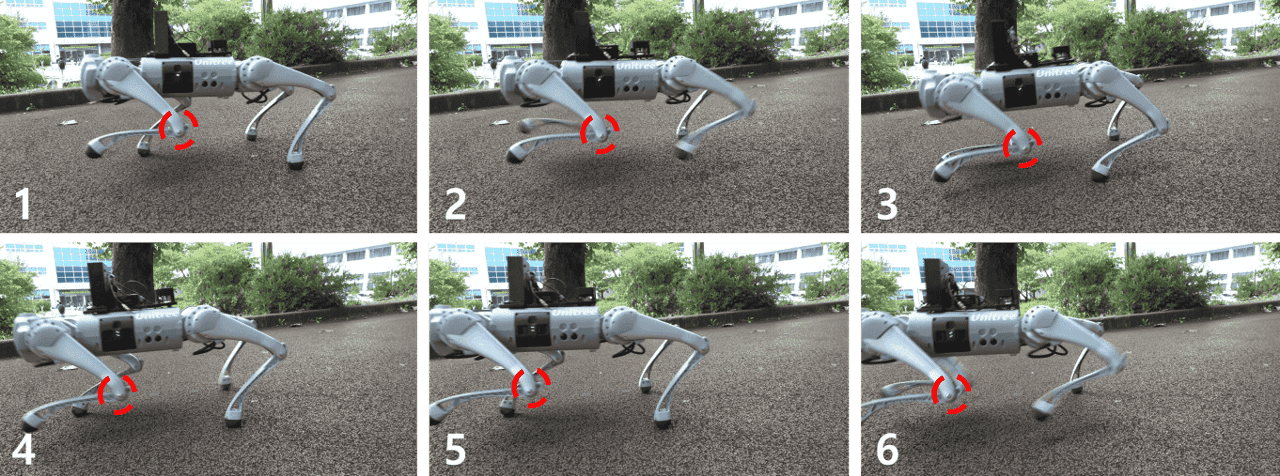} \vspace{-0.05in} \\  % 이미지 추가
{\footnotesize (c) Ours}
\end{tabular}

\caption{{\bf Quadrupedal locomotion with impaired joints.}
Our proposed framework aims to learn quadrupedal locomotion with impaired joints. 
In previous model~\cite{ref10}, if a joint is damaged by external or internal factors (a), it directly affects the locomotion ability and results in walking failure and robot body damage (b).
On the other hand, our proposed method (c) maintains stable locomotion capabilities even under impaired joint conditions (\ie, red dotted circles).
}
\vspace{-0.1in}
\label{fig:teaser} % 레이블 추가
\end{figure}

Furthermore, the impaired locomotion ability can cause significant damage to both robots and humans, shortening the lifespan of the robot.
As shown in~\figref{fig:teaser}-(a), if a joint is damaged by external or internal factors, it directly affects the locomotion ability and results in walking failure and robot body damage (\ie, \figref{fig:teaser}-(b)).
Therefore, it is necessary to maintain robust locomotion ability even with impaired joints to ensure the safety of the robot itself and humans.

However, previous quadrupedal locomotion studies have been focused on agile quadrupedal locomotion in rough terrains~\cite{ref5, ref20, ref7} rather than fault-tolerant locomotion. 
Also, the existing fault-tolerant locomotion study just provides a na\"ive fall-recovery functionality~\cite{ref6}, requires an accurate robot body model~\cite{ref21, ref23}, and not validated in real-world environments~\cite{ref24, ref25}.
Existing commercial robots also only provide basic fault detection or protective functions that shut down the system.

The main difficulty for fault-tolerant locomotion is that the locomotion strategies vary according to the damaged joint positions and status. 
For example, zero torque in the hip-roll position completely loses the whole Degree of Freedom (DoF) of one leg. 
However, motor locking up in the knee-pitch position may preserve most of the DoF of one leg.
It is especially challenging in model-based control~\cite{ref8, ref9}, such as Model Predictive Control (MPC) and Whole Body Control (WBC), which need an accurate body model and time-consuming model parameter tuning.
The model-based controls are highly limited in dealing with diverse impaired joint conditions.
Therefore, we need to cover the diverse malfunction scenarios of joint positions and status to achieve comprehensive locomotion ability with impaired joints.

To this end, we propose a novel deep reinforcement learning framework to enable a quadrupedal robot to walk with impaired joints. 
% We exploit deep reinforcement learning to learn comprehensive locomotion ability rather than utilizing model-based control, which needs case-by-case body models and specific model-parameter tuning for various impaired joint conditions.
The proposed framework contains three novel components: 1) a random joint masking strategy to provide diverse impaired joint scenarios, 2) a joint status estimator to judge each joint status and determine an appropriate locomotion strategy, and 3) progressive curriculum learning to make a single network conduct both normal gait and various joint-impaired gaits.
To evaluate the effectiveness of the proposed method, we conduct thorough and various experiments, including real-world demonstrations.
As shown in~\figref{fig:teaser}, our proposed framework enables the Unitree's Go1 robot to maintain robust locomotion capabilities even under various impaired joint conditions.

Our novelties can be summarized as follows:
\begin{enumerate}
\item
We propose a random joint masking strategy that simulates diverse joint malfunction scenarios, such as zero torque or locking up, by randomly masking joint actions and torques.
\item
We propose a joint status estimator that classifies whether each joint is normal or impaired based on previous observation history.
\item
We propose a progressive curriculum learning to allow a single network to conduct both normal gait and various joint-impaired gaits.
\end{enumerate}

\section{Related Works}
\label{sec:related works}

\subsection{Deep Reinforcement Learning for Quadrupedal Robot}
Deep reinforcement learning has been presented as an alternative direction for designing robot controllers without requiring explicit prior knowledge, such as dynamic models and inverse kinematics.
The recent literature has been focused on reducing training time~\cite{ref10}, agile quadrupedal locomotion~\cite{ref12,ref13}, and reducing sim-to-real gap~\cite{ref14,ref15,ref17}. 
Specifically, Rudin~\etal~\cite{ref10} greatly shortens the entire training time by utilizing massive parallel GPU processing. 
A number of study makes the agent robustly or quickly walk in challenging terrains, such as high-speed running extensions \cite{ref12} and fast locomotion on deformable terrains \cite{ref13}.
Another branch is reducing the domain gap between simulation and the real world by adapting environmental state~\cite{ref11,ref16}, using domain randomization~\cite{ref17,ref18,ref19}, and leveraging teach-student training method~\cite{ref20,ref11,ref12}.

However, all previous works assume the ideal condition that every joint is fully functional. 
In the real world, it is often violated by external accidents and internal hardware issues and leads to a complete loss of robot locomotion functionality.
Therefore, this paper focuses on learning robust quadruped locomotion under an impaired joint condition, which has been less explored despite its importance.

\subsection{Fault Tolerance Control for Quadrupedal Locomotion}

Most studies about fault-tolerant quadrupedal locomotion have been proposed in model-based control systems~\cite {ref21,ref22,ref23}.
They usually aimed to solve a single joint locking case with inverse kinematics~\cite{ref21}, gait modeling~\cite{ref22}, and whole-body control~\cite{ref23}. 
However, these methods often relied on a specific robot dynamic model, case-by-case models for various failure cases, and time-consuming model parameter tuning. 
A few fault tolerance algorithms utilizing deep reinforcement learning have been proposed~\cite{ref24,ref25}. 
However, the proposed methods were only evaluated in a simulation environment and are not publicly available.

Unlike the previous method, our proposed method handles various joint failure cases, including zero joint torque and joint locking with varying joint positions. 
Furthermore, the proposed method is evaluated in both simulation and real-world indoor and outdoor environments to verify its effectiveness.

\section{Proposed Learning Strategy}
\label{sec:method}

\subsection{Overview of framework}

\begin{figure*}[t]
\begin{center}
{
\begin{tabular}{c}
\includegraphics[width=0.9\linewidth]{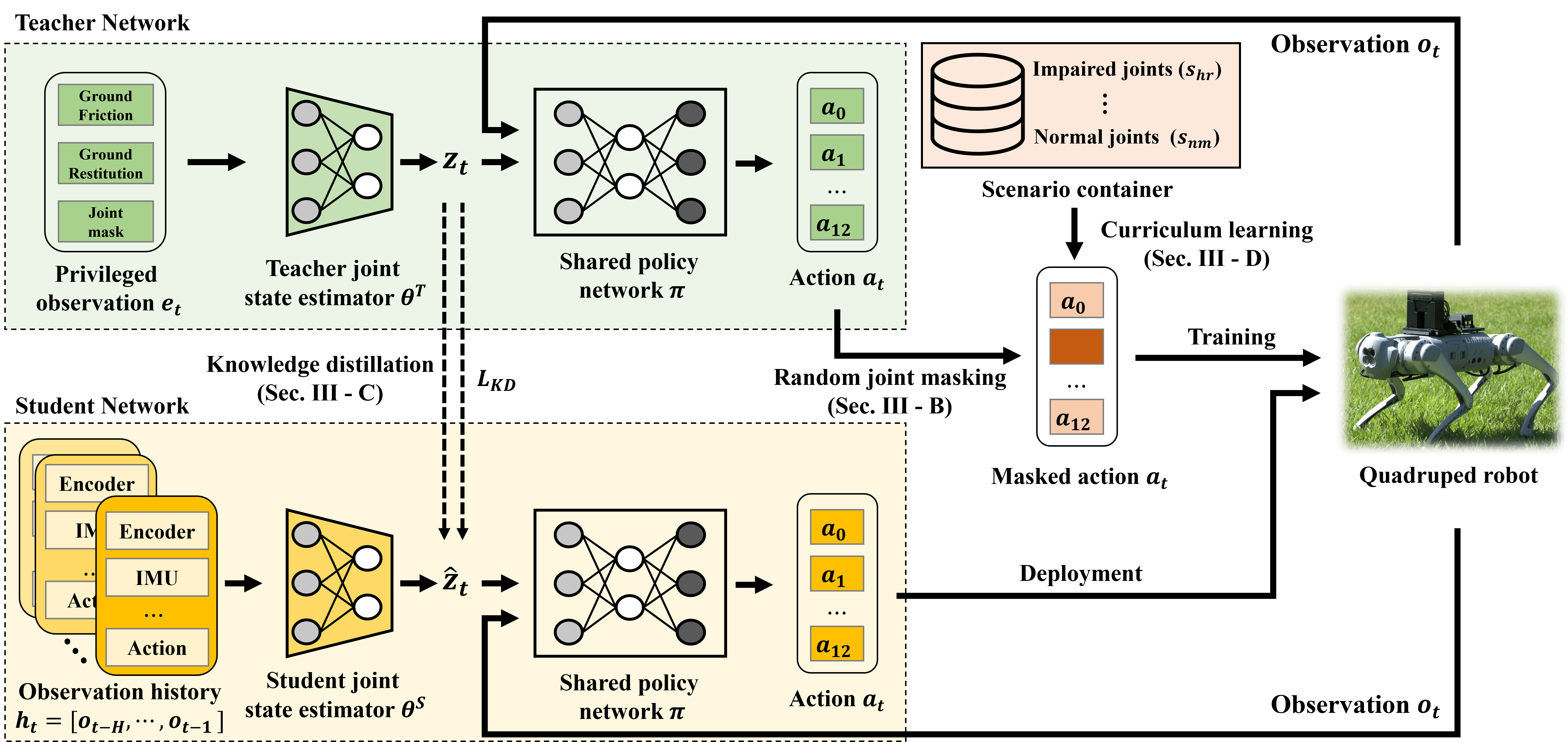} \\

\end{tabular}
}
\end{center}
%\vspace{-0.1in}
\caption{{\bf Overall pipeline of our proposed training framework.} 
Our proposed method aims to learn quadrupedal locomotion with impaired locomotion. For this purpose, we first enforce impaired joint scenarios to the robot by randomly masking joint action $a_t$ (\secref{sec:RJM}). Also, to make the robot estimate the current joint status (\ie, whether all joints are available or partially impaired) through observation history, we perform imitation learning between teacher and student latent vectors $\hat{z}_t$ and $z_t$ (\secref{sec:JSE}). Additionally, we propose a progressive curriculum learning strategy to enable comprehensive locomotion ability for a normal joint and various impaired joint conditions (\secref{sec:PCL}). Based on the proposed components, the robot is able to walk in not only a normal joint condition but also various impaired joint conditions.
}
\label{fig:teacher}
\vspace{-0.1in}
\end{figure*}

Our proposed framework consists of three components, as shown in~\figref{fig:teacher}; random joint masking to simulate diverse impaired joint conditions~(\secref{sec:RJM}), joint status estimator $\theta^S$ to judge the current joint status~(\secref{sec:JSE}), and progressive curriculum learning~(\secref{sec:PCL}) to allow a single policy network $\pi$ to conduct diverse normal and impaired quadrupedal locomotion.

\textbf{Architecture.}
Our framework contains teacher-student joint status estimators (\ie, $\theta^T$ and $\theta^S$) and shared policy network $\pi$.
Our proposed framework utilizes teacher-student knowledge distillation~\cite{hinton2015distilling} and privileged observation~\cite{ref11} to train a student joint status estimator.
The policy network, which is shared between teacher and student networks, is trained through Proximal Policy Optimization (PPO)~\cite{schulman2017proximal}.

\textbf{Training data acquisition.} 
Given privileged observation\footnote{Our privileged observation includes ground friction, ground restitution, and joint mask value for all joint status (\ie, 0 is normal, 1 to 12 indicates each impaired joint condition), defined as $e_t=[fr_t, gr_t, m_t]^T$, where $fr_t \in \mathbb{R}^{1}$, $gr_t \in \mathbb{R}^{1}$, and $m_t \in \mathbb{R}^{1}$} $e_t$, the teacher joint status estimator $\theta^{T}$ embeds the current environmental information as latent vector $z_{t}$. 
The latent $z_{t}$ and observation\footnote{Please refer to Appendix for observation and action details.} $o_{t}$ fed into policy network $\pi$.
After that, the policy network predicts the desired action\footnotemark[2] $a_t$ for all joints.
At this moment, the actions and joint torques are randomly masked (\ie, disabled) based on the current curriculum scenario.
The masked action forces the robot to move without utilizing the masked joint.
This process is iterated to make training data within the current curriculum scenario.
As a result, the policy network is trained to predict appropriate locomotion behavior according to the current joint status.

\begin{table}[t]
    \centering
    \caption{\textbf{Selectively utilized reward terms}. 
    The three reward terms promote normal gait locomotion (\eg, The values ($C_{foot}^{cmd}$, $s_{y}^{cmd}$) are generated from gait algorithm~\cite{margolis2023walk}). We exclude the three terms for impaired joint scenarios.
    Please refer to these works~\cite{ref10, ref12, margolis2023walk} for full reward table and descriptions.}
    \vspace{-0.1in}
    \begin{center}
    \resizebox{0.99\linewidth}{!}
    {
        \def\arraystretch{1.5}
        \footnotesize
        \begin{tabular}{c|c|c} 
        \toprule
        Reward                                                         & Equation                   & weight                       \\ \hline
        { Swing phase tracking}           & $\sum [1 - C_{\text{foot}}^{\text{cmd}}(\theta^{\text{cmd}}, t)] \exp\left(-|f^{\text{foot}}|^2/0.25\right)$ & 4.0   \\ \hline
        { Stance phase tracking}       & $\sum [C_{\text{foot}}^{\text{cmd}}(\theta^{\text{cmd}}, t)] \exp\left(-|V_{xy}^{\text{foot}}|^2/0.25\right)$ & 4.0   \\ \hline
        { Foot swing tracking} & $(p_{x,y,foot}^f - p_{x,y,foot}^{f, \text{cmd}}(s_y^{\text{cmd}}))^2$
       & -10.0 \\
        \bottomrule
        \end{tabular}

    }
    \end{center}
    \label{tab:Reward_table}
    \vspace{-0.1in}
\end{table}

\textbf{Reward.} 
The reward is calculated after conducting the masked action.
We utilized the same reward function used in the previous studies~\cite{ref10, ref12, margolis2023walk}. 
However, we found some reward terms that promote normal gait locomotion are ineffective for learning impaired quadrupedal locomotion. 
Therefore, we exclude the three reward terms (\ie,\cref{tab:Reward_table}) for impaired joint scenarios. 
However, we still include the three terms for normal joint scenarios.
Based on the training data acquisition process and reward terms, the policy network $\pi$ is optimized using PPO algorithm~\cite{schulman2017proximal} to maximize total expected reward~\eqnref{equ:teacher_objective}.
\begin{equation}
\label{equ:teacher_objective}
    J(\pi) = \mathbb{E}_{r \sim p(r|\pi)} \left[ \sum_{t=0}^{T-1} \gamma^t r_t \right],
\end{equation}
where $\gamma$ is the discount factor, $r_{t}$ is the weighted sum reward term at the time $t$, and $T$ is the scenario length.

\subsection{Random Joint Masking Strategy}
\label{sec:RJM}

Learning quadrupedal locomotion with impaired joints in the real world is a challenging problem due to various hardware and software issues, such as the difficult data acquisition process, potential severe damage to the robot body, non-observable information, and inaccurate robot state.
Therefore, simulating impaired joint conditions is a safe and reliable solution to learning quadrupedal locomotion with impaired joints.
To this end, we propose a random joint masking strategy to simulate diverse and plentiful joint malfunction scenarios, such as zero torque and locking up, at various joint positions.
During the training process, the proposed strategy randomly sets joint torque as zero (\ie, the joint is not functional anymore) and also sets the predicted action value of masked joint position as zero (\ie, force not to use the deactivated joint).
The detailed process is as follows.

\subsubsection{Joint malfunction scenario assignment}
Based on the current curriculum learning level, the current scenario assigns a normal joint condition or various joint malfunction scenarios (\eg, malfunction in front-left-hip-pitch joint).
If the current scenario is a normal joint condition, the predicted action $a_t$ is directly delivered to the robot.
If not, the predicted action and torque at the assigned joint position are masked out. 
Note that, during the training process, the multiple agents simultaneously conduct various normal and malfunction scenarios~\cite{ref10}.

\subsubsection{Joint action and torque masking}
When the joint malfunction position is assigned, the torque $\tau_{\text{des}}$ and action value $a_t$ of the assigned position are set to zero, as shown in~\eqnref{eqn:torque_masking} and~\eqnref{eqn:action_masking}.
\begin{equation} 
    \tau_{\text{des}}(p) \leftarrow \begin{cases}
    0 & \text{if impaired joint} \\
    \tau_{\text{des}}(p) & \text{otherwise}
\end{cases},
\label{eqn:torque_masking}
\end{equation}
where $p$ indicates joint position.
Making the desired torque zero means the joint is not functional.
\begin{equation} 
    a_{t}(p) \leftarrow \begin{cases}
    0 & \text{if impaired joint} \\
    a_{t}(p) & \text{otherwise}
\label{eqn:action_masking}
\end{cases}
\end{equation}
Also, masking the predicted action value as zero forces the agent not to use the deactivate joint.
Therefore, the proposed strategy provides diverse and plentiful experiences in which various joints are not functional and won't move.
As a result, the agent needs to learn a locomotion strategy that doesn't use the impaired joint.

\subsection{Joint Status Estimator}
\label{sec:JSE}
Another important requirement for impaired quadruped locomotion is to identify the current joint status, whether the joints are fully functional or damaged, based on the observable information in the real world.
This enables the agent to determine an appropriate locomotion strategy according to the current joint status.
For this purpose, we utilize teacher-student knowledge distillation~\cite{hinton2015distilling} and the idea of privileged observation~\cite{ref20, ref11, ref12}.
The previous works~\cite{ref20, ref11, ref12} estimate embedded states of environmental variables that are challenging to know in the real world, such as ground friction and restitution, based on past observation history.

By extending this idea, we propose a joint status estimator $\theta^S$ that can estimate implicit representation of the current joint functionality and environmental variables based on the past observation history.
The joint status estimator is trained with knowledge distillation from teacher estimator $\theta^T$ to student estimator $\theta^S$.
\subsubsection{Teacher joint status estimator} 
We fed the joint mask (ranging from 0 to 12) along with environmental variables to the teacher estimator as a privileged observation $e_t$.
The joint mask indicates that all joints are fully functional or a certain joint is malfunctioning.
Then, the teacher network implicitly represents the privileged observation as latent vector $z_t$.
\subsubsection{Student joint status estimator} 
On the other hand, student estimators cannot leverage privileged observation in the real world.
Therefore, the student estimator aims to estimate the same latent vector of the teacher network based on available observation information in the real world (\ie, $[o_{t-H}, ..., o_{t-1}]$, where $H$ is set to 30).
To this end, the student estimator is trained to minimize the difference between the teacher latent vector $z_t$ and student latent vector $\hat{z_t}$, as shown in ~\eqnref{Equ:Student}.
This knowledge distillation loss $L_{KD}$ enables the student estimator to estimate the implicit representation for the current joint status and environmental variable information with a given past observation history. 
\begin{equation}
    L_{KD}(\hat{z_t}, z_t) = \lVert \hat{z_t} - z_t \rVert^2
\label{Equ:Student}
\end{equation}

\subsection{Progressive Curriculum Learning for Unified Policy}
\label{sec:PCL}
% ~\tabref{tab:test}
\setlength{\textfloatsep}{2pt}% Remove \textfloatsep
\begin{algorithm}[t]
\caption{Progressive Curriculum Learning}
\label{algo:training}
\SetAlgoLined
 Initialize policy network $\pi$, status estimators $\theta^T$, $\theta^S$; \\
 initialize curriculum scenario container $C$ = [$s_{nm}$]; \\ 
 Empty buffers $D_1$, $D_2$; \\
 \For{$0 \leq itr \leq N_{\mathrm{itr}}$}{ 
     \For{$0 \leq i \leq N_{\mathrm{env}}$}{ 
        $s_i \gets AssignSenario(C)$;   \textcolor{gray}{\# Sec. \uppercase\expandafter{\romannumeral3}-B}      \\
        \For{$0 \leq t \leq T$}{
            $z_t \gets \theta^T(e_t)$;,  $\hat{z_t} \gets \theta^S(h_t)$;   \\
            $a_t \gets \pi(o_t, z_t)$;              \\
            $a_t \gets Masking(a_t, s_i)$;   \textcolor{gray}{\# Sec. \uppercase\expandafter{\romannumeral3}-B}      \\
            $o_{t+1}, e_{t+1}, r_t \gets $ envs[$i$].step($a_t$); \\
            Store $(o_t, e_t, a_t, r_t)$, $(\hat{z_t}, z_t)$ in $D_1$, $D_2$; \\
        }
 
     }
    Update $\pi^T$ and $\theta^T$ using PPO \cite{schulman2017proximal};  \\
    Update $\theta^S$ with $L_{KD}$; \textcolor{gray}{ \# Sec. \uppercase\expandafter{\romannumeral3}-C} \\
    Empty buffers $D_1$, $D_2$; \\
    \textcolor{gray}{\# Sec. \uppercase\expandafter{\romannumeral3}-D, Progressive curriculum learning} \\
    $R_{avg} \gets \sum_{i=0}^{N_{env}}J(\pi)/N_{env}$; \\
    \If{$R_{avg} > th_{\mathrm{level1}}$} 
    {
        $C$ = [$s_{nm}$, $s_{kp}$];  \\
        \If{$R_{avg} > th_{\mathrm{level2}}$}
        {
            $C$ = [$s_{nm}$, $s_{kp}$, $s_{hp}$];  \\
            \If{$R_{avg} > th_{\mathrm{level3}}$}
            {
                $C$ = [$s_{nm}$, $s_{kp}$, $s_{hp}$, $s_{hr}$];  
            }
        }
    }
     
}
\end{algorithm}

Locomotion strategies may vary depending on the impaired joint position and status.
Training each network for each locomotion strategy greatly increases memory requirement and reduces the flexibility and scalability of the network.  
However, training a single network to handle various locomotion strategies is also a challenging problem because of the catastrophic forgetting~\cite{atkinson2021pseudo, kirkpatrick2017overcoming, kaushik2021understanding}.
In other words, the network forgets previously trained locomotion strategy while only preserving the newly learned strategy.
Therefore, it leads to imbalanced performance for the various impaired joint scenarios.

To resolve this issue and make a single network encompass both normal and various joint-impaired locomotion capabilities, we propose a progressive curriculum learning method. 
The proposed method progressively includes more difficult joint-impaired conditions while preserving overall performance over the current curriculum learning level.
As shown in~\cref{algo:training}, the initial curriculum level only includes a normal joint condition $s_{nm}$, the easiest and basic locomotion.
% Also, we believe that training normal locomotion at first would be helpful to learn joint-impaired locomotion.
After that, whenever the reward mean value $R_{avg}$, averaged over the current curriculum level, satisfies certain criteria, the curriculum learning level gradually includes more difficult impaired joint conditions in the order of knee-pitch, hip-pitch, and hip-roll. 
If the curriculum level reaches the final stage, all joint conditions (\ie, normal and abnormal joints) are equally assigned to the scenarios.% (\ie, the ratio of 1/13).

Note that the order is decided by the disturbance level based on empirical findings.
When one of the legs is impaired because of a joint malfunction, the impaired leg can be considered a disturbance factor to the other three legs.
In this case, we found the most significant disturbance occurs when the torque of the Hip-Roll joint, which is closest to the robot's body, becomes zero.
The tendency of disturbance decreases as the impaired joint is far from the robot's body (\ie, knee-pitch is the least disturbance).
Therefore, we proceeded with the curriculum learning level in the order of knee-pitch, hip-pitch, and hip-roll.

\begin{figure*}[t]
\centering % 중앙 정렬
% \begin{tabular}{c}
\begin{tabular}{c@{\hskip 0.001\linewidth}c@{\hskip 0.001\linewidth}c}
\includegraphics[width=0.3\linewidth]{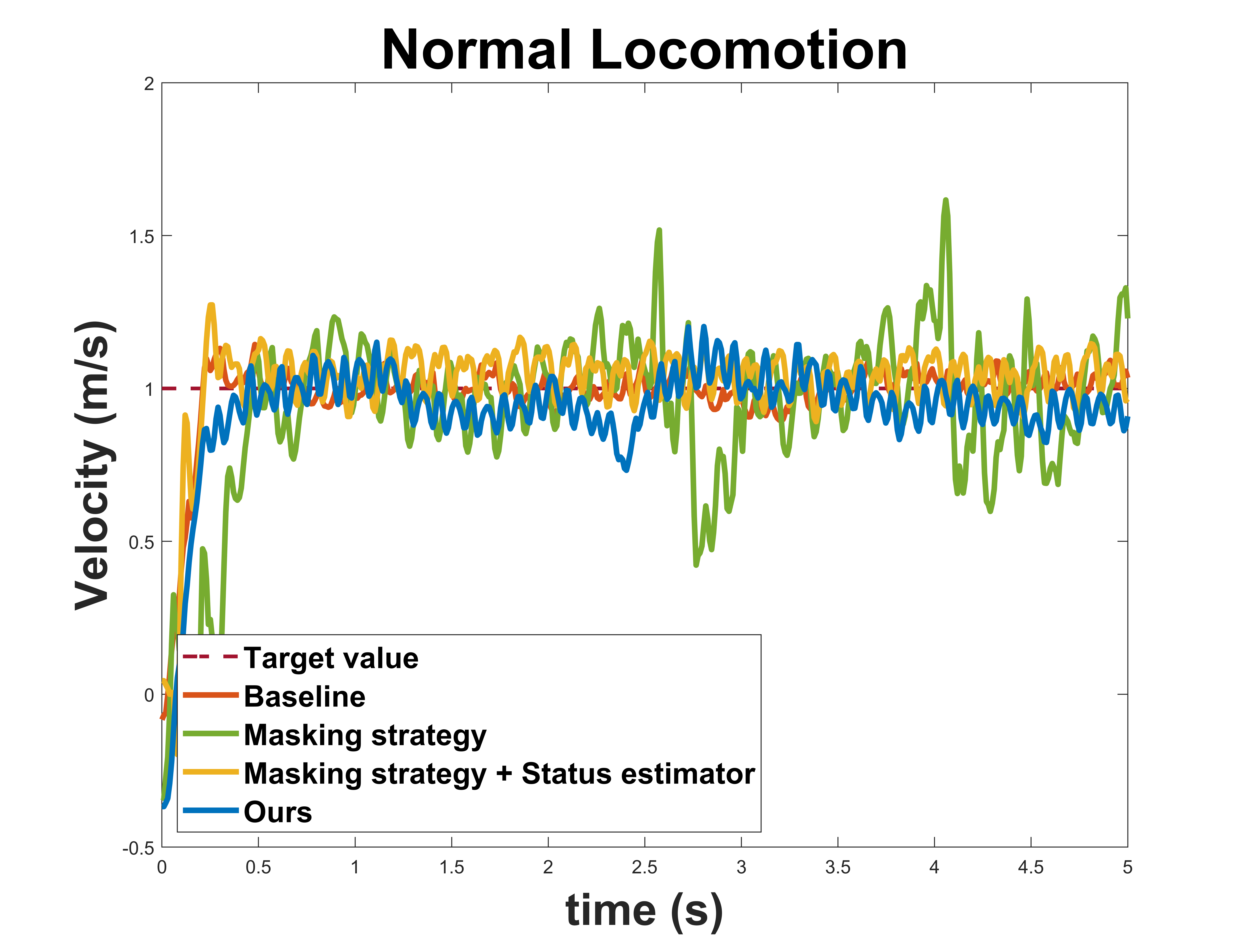} & % 이미지 추가
\includegraphics[width=0.3\linewidth]{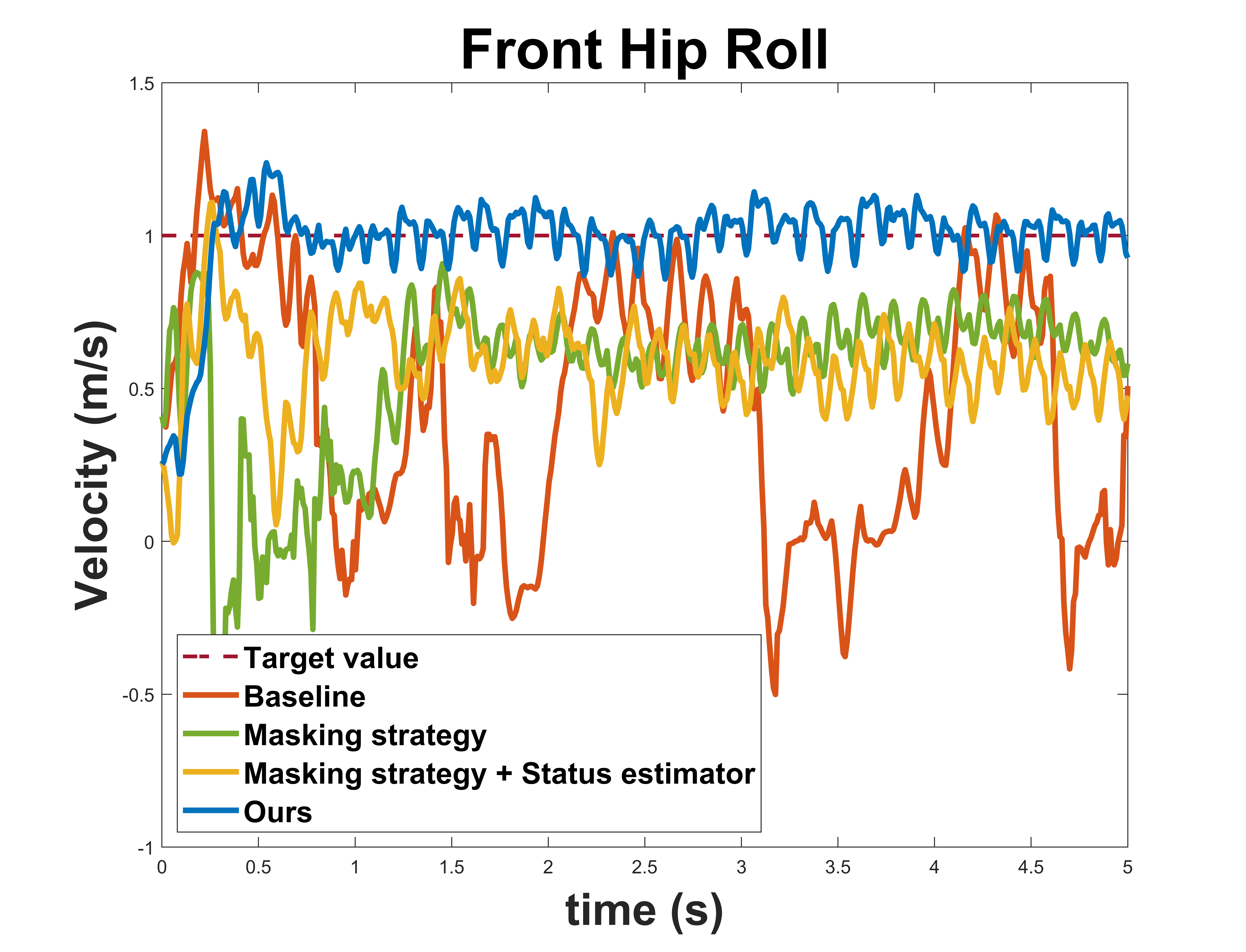} & % 이미지 추가
\includegraphics[width=0.3\linewidth]{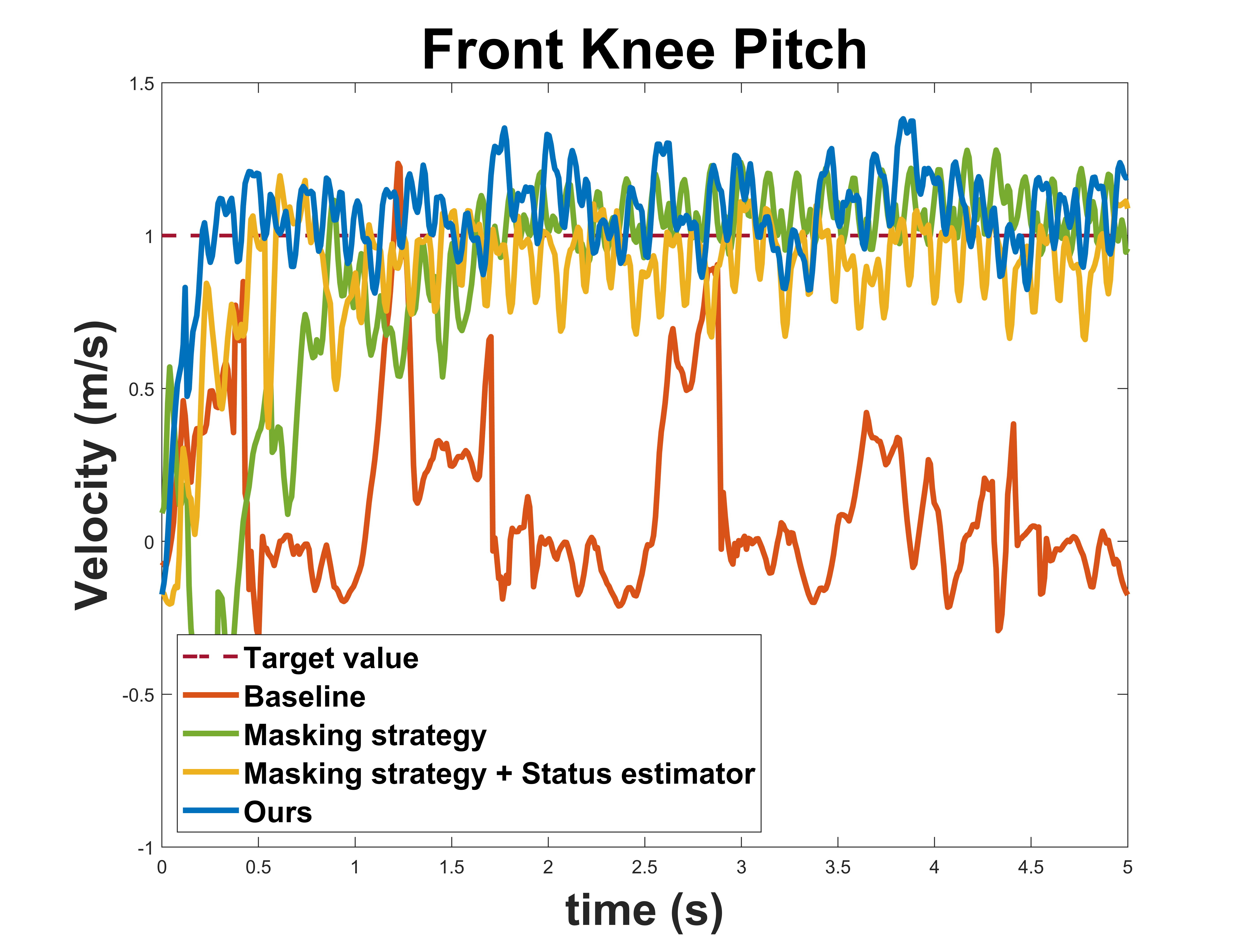} \\ % 이미지 추가
{\footnotesize (a) Normal joint condition} &  {\footnotesize (b) Zero-torque (Hip-roll)} & {\footnotesize (c) Zeor-torque (Knee-pitch)}\\
\end{tabular}
% \vspace{-0.1in} % 캡션과 그림 사이 간격 조정
\caption{{\bf Response graph comparison for body velocity command in Isaac Gym simulation~\cite{makoviychuk2021isaac}.}
We compare baseline~\cite{ref10} and variants of the proposed method (\ie, an ablation study of each component).
Given the forward body velocity command (\ie, 1.0 m/s), all methods show similar performance in a normal joint condition (a).
However, when a joint has malfunctioned (\ie, zero-torque) at a randomly joint (b,c), our proposed framework immediately responds and shows stable tracking ability.
}

\label{fig:Sim} % 레이블 추가
\vspace{-0.15in}

\end{figure*}

\begin{table}[t]
\centering
\caption{\textbf{Tracking error comparison in Issac Gym simulation}. 
Given the forward velocity command, we compare the tracking error of each method with the Root-Mean-Square-Error (RMSE) metric.
The best and runner-up performances in each block are highlighted in \textbf{bold} and \underline{underline}.
}
\vspace{-0.1in}
\begin{center}
\resizebox{1.0\linewidth}{!}
{
    \def\arraystretch{1.05}
    \footnotesize
    \begin{tabular}{c|c|c|c|c|c|c} 
    \toprule
    \multirow{3}{*}{Joint status} & \multicolumn{2}{c|}{\multirow{3}{*}{\shortstack[alignment]{Impaired \\ Joint position}}} & \multicolumn{4}{c}{\textbf{RMSE} $\downarrow$} \\ \cline{4-7}
    & \multicolumn{2}{c|}{} & \multirow{2}{*}{Baseline~\cite{ref10}} &  \multirow{2}{*}{+Masking} & +Masking &  \multirow{2}{*}{Ours} \\ 
    & \multicolumn{2}{c|}{} & & & +Status Est & \\ \hline
Normal      &\multicolumn{2}{c|}{None}        & \textbf{0.1469} & 0.2079 & 0.3009 & \underline{0.1699}     \\ \hline
\multirow{8}{*}{\shortstack[alignment]{Zero \\ Torque}} & \multirow{3}{*}{\shortstack[alignment]{Front\\ Left}} & HR   & 0.7089 & 0.5290 & 0.4364 & 0.1507\\ 
            &            & HP  & 0.9747 & 0.4899 & 0.2980 & 0.2668       \\
            &            & KP & 0.9252 & 0.4214 & 0.2580 & 0.1920       \\ 
            &\multicolumn{2}{c|}{\cellcolor{Gray1}Avg} & \cellcolor{Gray1}{0.8696} & \cellcolor{Gray1}{0.4801} & \cellcolor{Gray1}\underline{0.3308} & \cellcolor{Gray1}\textbf{0.2031} \\ \cline{2-7}
            & \multirow{3}{*}{\shortstack[alignment]{Rear\\ Right}} & HR   & 0.8928  & 0.2300 & 0.2475 & 0.1986        \\ 
            &            & HP  & 1.0320
     & 0.2584  & 0.1513     & 0.1959      \\
            &            & KP  & 0.8081
    & 0.3604   & 0.2003     & 0.1492       \\ 
            &\multicolumn{2}{c|}{\cellcolor{Gray1}Avg} & \cellcolor{Gray1}{0.9109} & \cellcolor{Gray1}{0.2829} & \cellcolor{Gray1}\underline{0.1997} & \cellcolor{Gray1}\textbf{0.1812} \\ \hline
\multirow{8}{*}{\shortstack[alignment]{Locked \\ Up}} & \multirow{3}{*}{\shortstack[alignment]{Front\\ Left}} & HR   & 0.2503 & 0.4803 & 0.2789 & 0.2003\\ 
            &            & HP  & 0.2456 & 0.5075 & 0.2896 & 0.2333    \\
            &            & KP & 0.2831  & 0.6684 & 0.2158 & 0.2456      \\ 
            &\multicolumn{2}{c|}{\cellcolor{Gray1}Avg} & \cellcolor{Gray1}\underline{0.2596} & \cellcolor{Gray1} {0.5520}& \cellcolor{Gray1}{0.2614} & \cellcolor{Gray1} \textbf{0.2264}\\ \cline{2-7}
            & \multirow{3}{*}{\shortstack[alignment]{Rear\\ Right}} & HR   & 0.6966  & 0.2704 & 0.1484 & 0.2001    \\ 
            &            & HP  & 0.7086
   & 0.3935   & 0.1405 & 0.1542     \\
            &            & KP & 0.7139   & 0.2614  & 0.2265 & 0.2298       \\ 
            &\multicolumn{2}{c|}{\cellcolor{Gray1}Avg} & \cellcolor{Gray1}{0.7063} & \cellcolor{Gray1}{0.3084}& \cellcolor{Gray1}\textbf{0.1718} & \cellcolor{Gray1}\underline{0.1947} \\ \hline
    \multicolumn{3}{c|}{\cellcolor{Gray1}Total Avg} & \cellcolor{Gray1}0.5787 & \cellcolor{Gray1}0.3662 & \cellcolor{Gray1}\underline{0.2529} & \cellcolor{Gray1}\textbf{0.1951} \\
    \bottomrule
    \end{tabular}

}

\end{center}
\label{tab:sim_table}
\end{table}

\section{Experimental results}
\label{sec:experiments}

\begin{figure*}[t]
\centering % 중앙 정렬
% \begin{tabular}{c}
\begin{tabular}{c@{\hskip 0.005\linewidth}c@{\hskip 0.005\linewidth}c@{\hskip 0.005\linewidth}c}
\includegraphics[width=0.23\linewidth]{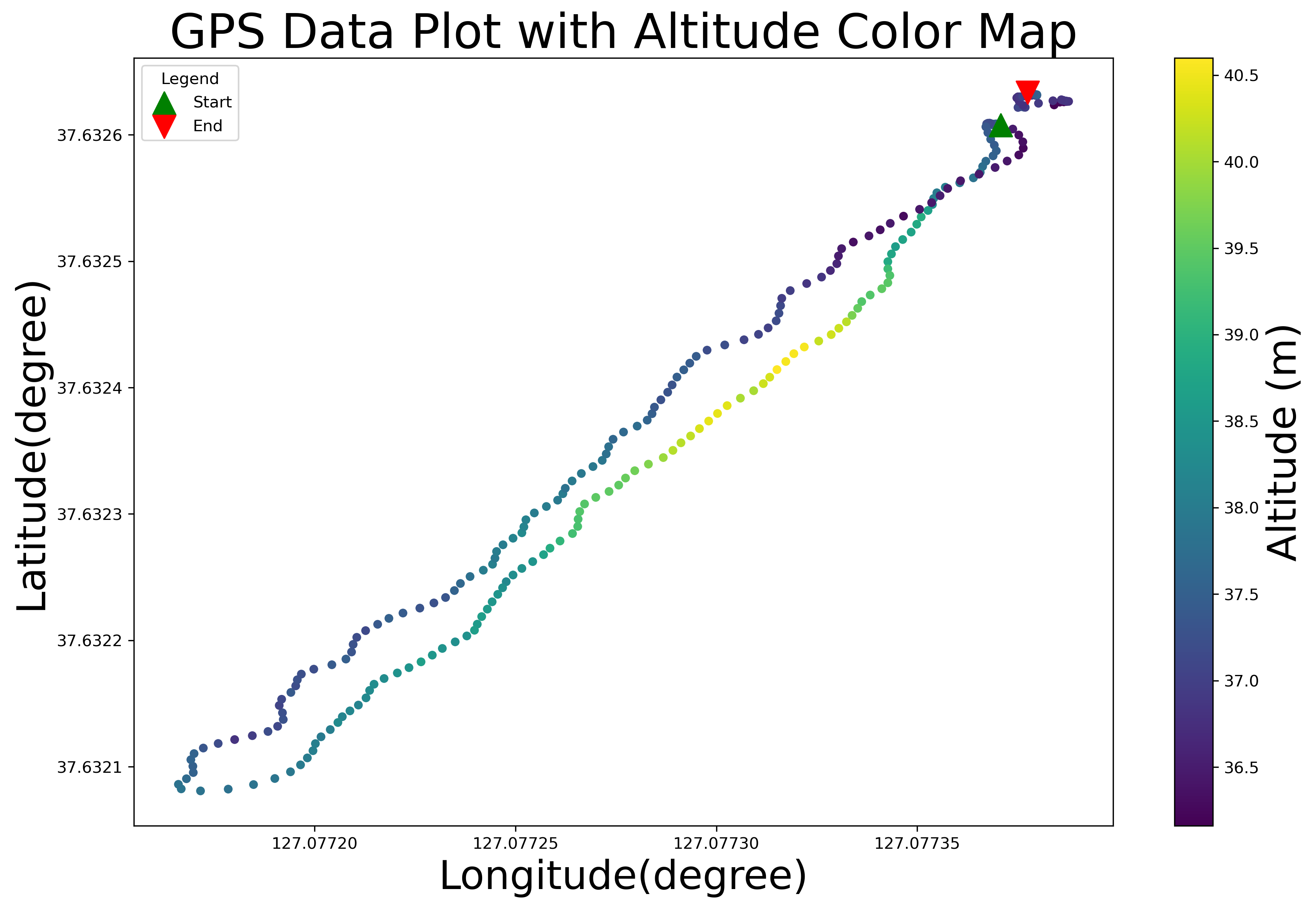} & % 이미지 추가
\includegraphics[width=0.23\linewidth]{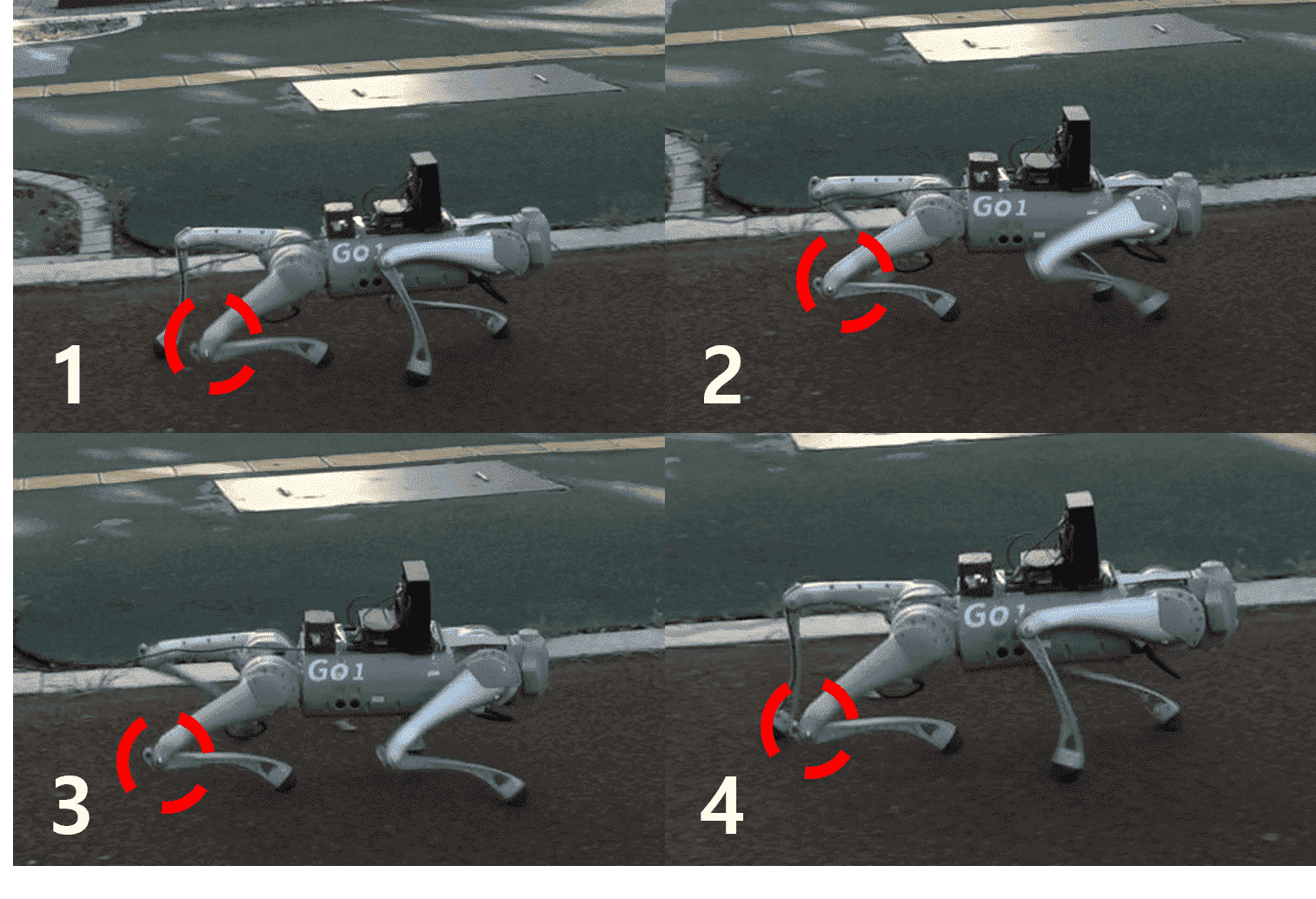} & % 이미지 추가
\includegraphics[width=0.23\linewidth]{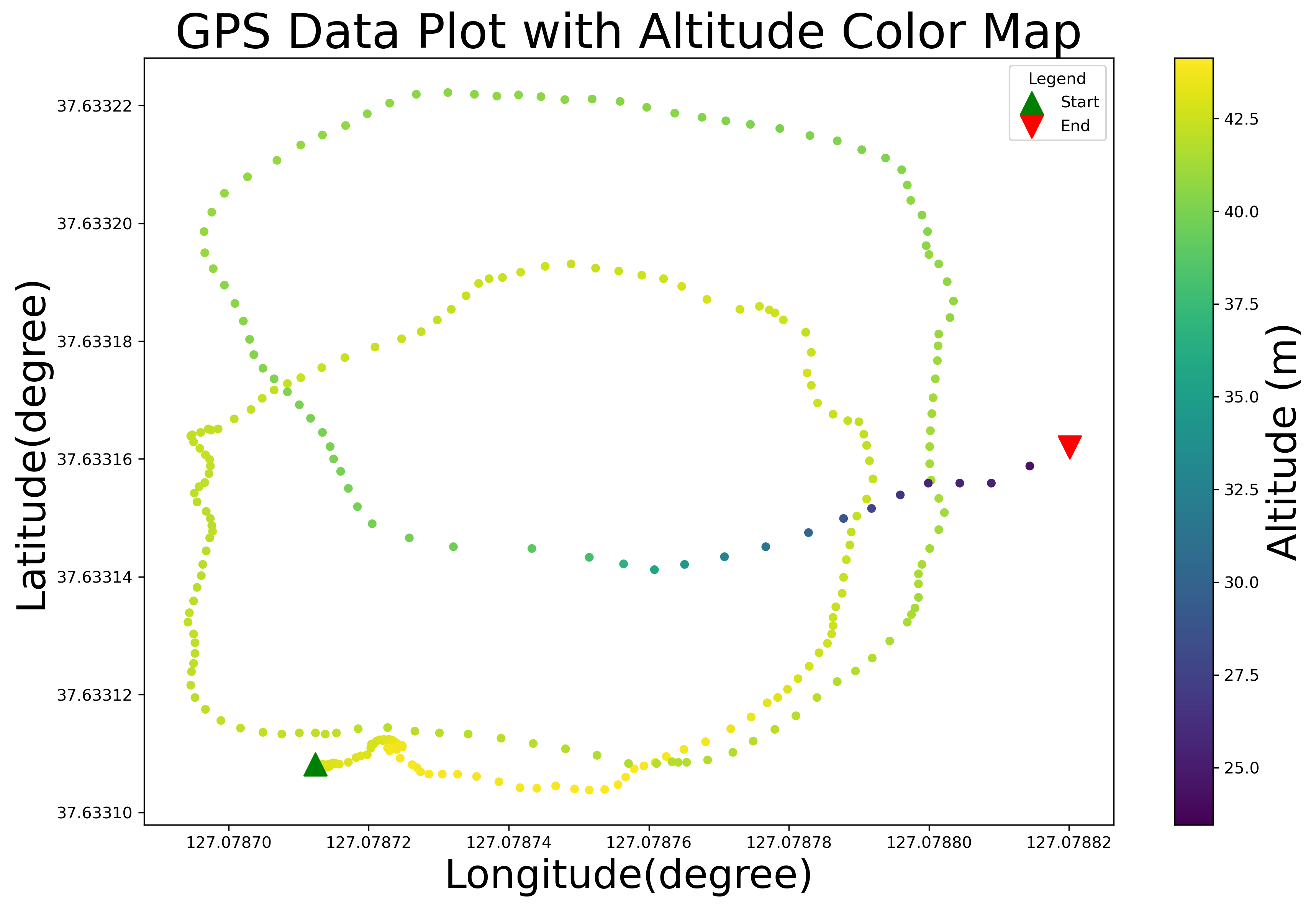} & % 이미지 추가

\includegraphics[width=0.23\linewidth]{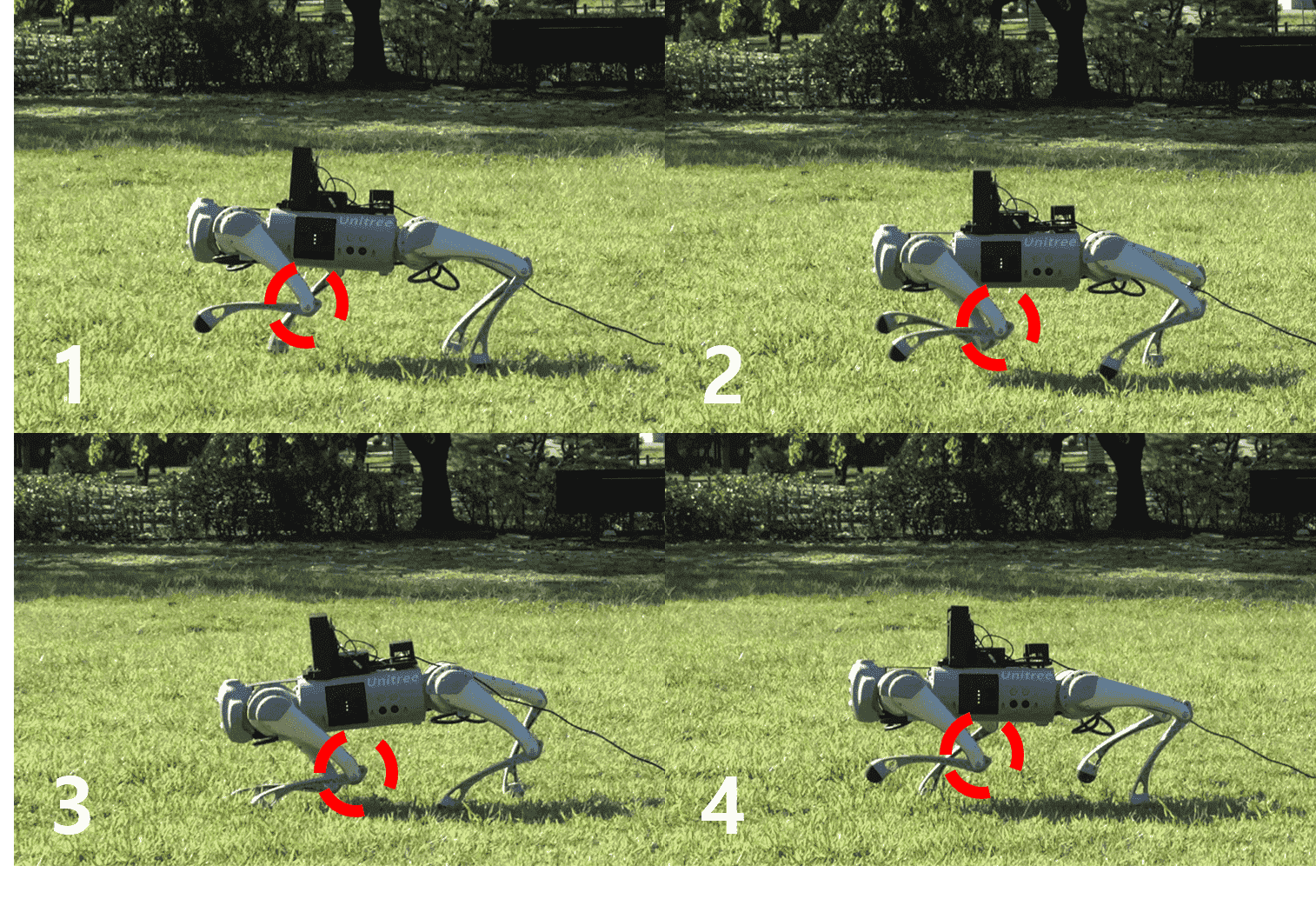} \\ % 이미지 추가
{\footnotesize (a) Course 1} &  {\footnotesize (b) Snapshot in Course 1} & {\footnotesize (c)  Course 2} & {\footnotesize (d) Snapshot in Course 2}\\
\end{tabular}
% \vspace{-0.1in} % 캡션과 그림 사이 간격 조정
\caption{{\bf Impaired quadrupedal locomotion in real-world outdoor environments.}
The trajectory was recorded using an RTK-GPS attached to the robot's body. 
Course 1 consists of gentle incline, decline, and level areas with clean and hard ground.
Course 2 is deformable ground covered with soft grass and small pebbles.}
\label{fig:outdoor} % 레이블 추가
\vspace{-0.1in} % 캡션과 그림 사이 간격 조정
\end{figure*}

\subsection{Quadrupedal Locomotion in Isaac Gym Simulation~\cite{makoviychuk2021isaac}}

We evaluate our proposed method with the baseline algorithm~\cite{ref10} and the variants of the proposed method.
Unfortunately, fault-tolerant reinforcement learning methods~\cite{ref24,ref25} are publicly unavailable, so we were not able to compare these methods.
The experimental results are shown in~\figref{fig:Sim} and ~\tabref{tab:sim_table}.
'Masking strategy' indicates we have added the proposed joint masking strategy (\ie, Sec \uppercase\expandafter{\romannumeral3}-B) to the Baseline~\cite{ref10} method.
'Masking strategy + status estimator' means we further have added joint status estimator (\ie, Sec \uppercase\expandafter{\romannumeral3}-C).
'Ours' is the final proposed framework.
For the evaluation, we gave the forward body velocity command of 1.0m/s and measured the tracking accuracy of each method using the Root-Mean-Square-Error (RMSE) metric.

All methods show similar performance in a normal joint condition (\figref{fig:Sim}-(a)).
However, when a certain joint has malfunctioned (\ie, zero-torque) randomly, baseline~\cite{ref10} loses the locomotion capability and doesn't follow the command at all.
On the other hand, our proposed framework immediately responds and shows stable and reliable tracking ability (\figref{fig:Sim}-(b,c)).
As shown in~\tabref{tab:sim_table}, our proposed method achieves the best and second-best performance over a normal joint and various impaired joint conditions.

The table also shows the effectiveness of each proposed component of our proposed framework.
Applying the random joint masking strategy to the baseline greatly decreased the tracking errors in impaired joint conditions.
However, the performance tendency was quite imbalanced according to each impaired joint scenario. 
This is because the agent didn't know about the current joint conditions, whether it is normal or impaired.
The joint status estimator resolved the issue and let the model know which joint was damaged or fully functional.
However, still, the performance tendency was imbalanced and the normal gait performance was decreased. 
Progressive curriculum learning resolved the imbalanced performance issue and boosted the overall locomotion ability over various joint conditions.
Based on the three proposed components, we finally built a single neural network to possess diverse quadrupedal locomotion capabilities in normal and various impaired joint conditions.

\subsection{Empirical Analysis of Impaired Quadrupedal Locomotion in Real-world Laboratory Environments}

In this section, we empirically analyze the performance of impaired quadrupedal locomotion in a real-world indoor environment by using a motion capture system. 
We applied our learned model to the Unitree's Go1 robot without any fine-tuning stage. 
For the experiments, we give joint failure scenarios to the robot by forcing the joint at a fixed state (\ie, locked up) or setting the joint torque as zero (\ie, zero-torque).
Also, we assign forward-moving and rotation-in-place tasks in normal joint and joint malfunction conditions. 
We measure the robot's body and yaw velocity using six OptiTrack motion capture cameras to evaluate the performance of forward-moving and in-place rotation ability.
The body and yaw velocities are averaged over a 5m distance and 10 seconds.

The experimental results are shown in~\cref{tab:real}.
The Unitree's Go1 robot has reported 1.35m/s body velocity and 3.46 rad/s yaw velocity in a normal joint condition.
Impaired locomotion generally shows slightly worse but comparable results compared to normal quadrupedal locomotion.
Compared to the baseline model~\cite{ref10} that was tumbled down under impaired joint conditions (\ie, \figref{fig:teaser}-(b)), we believe our proposed framework significantly increases the robot's stability and reliability regarding locomotion capability.
Also, our proposed framework sustains the forward-moving and rotation tasks against diverse impaired joint positions (\eg, hip-roll, hip-pitch, and knee-pitch).
Still, there is a lot of room to improve in that the performance of impaired locomotion decreases according to the impaired joint positions and front-biased center of gravity.
For example, malfunction at the hip-pitch position much degenerates locomotion capability than the keen-pitch joint, due to the disturbance level that is proportional to the distance to the robot body.
%Also, malfunction at the front leg with torque zero condition causes degeneration of the rotation performance, which might be due to the front-biased center of gravity.
We will investigate this problem in future work.

\begin{table}[t]
\centering
\caption{\textbf{Performance comparison of impaired quadrupedal locomotion using motion capture data}. 
}
\vspace{-0.1in}
\begin{center}
\resizebox{1.0\linewidth}{!}
{
    \def\arraystretch{1.0}
    \footnotesize
    \begin{tabular}{c|c|c|cc} 
    \toprule
    \multirow{2}{*}{Joint status} & \multicolumn{2}{c|}{\multirow{2}{*}{\shortstack[alignment]{Impaired \\ Joint position}}} & \multicolumn{2}{c}{Motion capture data} \\ \cline{4-5}
            &   \multicolumn{2}{c|}{}  & Body velocity(m/s)  & Body Yaw(rad/s) \\ \hline
Normal       & \multicolumn{2}{c|}{None}     & \textbf{1.3536}              & \textbf{3.4643}     \\ \hline
\multirow{8}{*}{Zero-Torque} & \multirow{3}{*}{Front Left} & HR   & 0.9964              & 2.6009     \\ 
            &            & HP  & 1.1506              & 2.6628     \\
            &            & KP & 1.2687              & 2.7093     \\ 
            &\multicolumn{2}{c|}{\cellcolor{Gray1}Avg} & \cellcolor{Gray1}\textbf{1.1386} & \cellcolor{Gray1}\textbf{2.6577} \\ \cline{2-5}
            & \multirow{3}{*}{Rear Right} & HR   & 1.044               & 3.3191     \\ 
            &            & HP  & 1.3121              & 3.532      \\
            &            & KP & 1.3171              & 3.0147     \\ 
            &\multicolumn{2}{c|}{\cellcolor{Gray1}Avg} & \cellcolor{Gray1}\textbf{1.2244} & \cellcolor{Gray1}\textbf{3.2886} \\ \hline
\multirow{8}{*}{Locked up} & \multirow{3}{*}{Front Left} & HR   & 0.9702              & 3.2931     \\ 
            &            & HP  & 1.0050              & 3.7662     \\
            &            & KP & 1.0580              & 3.9492     \\ 
            &\multicolumn{2}{c|}{\cellcolor{Gray1}Avg} & \cellcolor{Gray1}\textbf{1.0111} & \cellcolor{Gray1}\textbf{3.6695} \\ \cline{2-5}
            & \multirow{3}{*}{Rear Right} & HR   & 1.0682               & 3.4407     \\ 
            &            & HP  & 1.0416              & 3.6306      \\
            &            & KP & 1.1703              & 3.6719     \\ 
            &\multicolumn{2}{c|}{\cellcolor{Gray1}Avg} & \cellcolor{Gray1}\textbf{1.0934} & \cellcolor{Gray1}\textbf{3.5811} \\ \hline
    \bottomrule
    \end{tabular}

}

\end{center}
\label{tab:real}
\end{table}

\subsection{Impaired Quadrupedal Locomotion in Real-world Outdoor Environments}

We further evaluated that our proposed framework can walk with impaired joints in real-world outdoor environments.
To evaluate performance in outdoor environments, we selected two different courses.
Course 1 consists of gentle incline slopes, decline slopes, and level areas with clean and hard grounds.
Course 2 is almost a flat region but the terrain is deformable ground covered with soft grass and small pebbles.
We measured the robot's trajectory using Real-Time Kinematic (RTK) GPS attached to the robot's body.
The outdoor evaluation results are shown in~\figref{fig:outdoor}.
We can observe that the agent robustly walks in various terrains even under impaired joint conditions.
The agent traverses two different terrains with a total distance of 0.5 km based on learned impaired quadrupedal locomotion strategies (\ie, \figref{fig:outdoor}-(b,d)).
Further results can be found in the supplementary video.
In future work, we plan to study walking in challenging terrains such as mountains and caves.

\section{Conclusion}

In this paper, we proposed a novel deep reinforcement learning framework to enable a quadrupedal robot to walk with impaired joints.
The proposed framework contains three novel components: 1) a random joint masking strategy to provide diverse impaired joint scenarios, 2) a joint status estimator to judge each joint status and determine an appropriate locomotion strategy, and 3) progressive curriculum learning to make a single network conduct both normal gait and various joint-impaired gaits.
The proposed framework is thoroughly verified in simulation environments, real-world indoor environment using motion capture, and real-world outdoor environments. 
As a result, our proposed framework enables the Unitree's Go1 robot to maintain stable and reliable locomotion capability even under various impaired joint conditions.

%%%%%%%%%%%%%%%%%%%%%%%%%%%%%%%%%%%%%%%%%%%%%%%%%%%%%%%%%%%%%%%%%%%%%%%%%

\section*{ACKNOWLEDGMENT}

This work was supported by the Ministry of Trade, Industry and Energy (MOTIE, Korea) under the Industrial Technology Innovation Program. Grant No. 20026194, "Development of Human-Life Detection and Fire-Suppression Solutions based on Quadruped Robots for Firefighting and Demonstration of Firefighting Robots and Sensors".

\section*{Appendix}
\textbf{Observation.}
Observation $o_t\in\mathbb{R}^{62}$ consists of joint angles $q_t \in \mathbb{R}^{12}$, joint angular velocities $\dot{q}_t \in \mathbb{R}^{12}$, gravity vector $g_t\in\mathbb{R}^{3}$, foot touchdown location $f_t \in\mathbb{R}^{4}$, x and y velocity, yaw angular velocity, body height, foot swing frequency, body pitch and body roll $c_t\in\mathbb{R}^{7}$, current action $a_t\in\mathbb{R}^{12}$, and previous action $a_{t-1}\in\mathbb{R}^{12}$.

\textbf{Action.}
Action $a_t$ are defined as the angles of each joint, $a_t\in\mathbb{R}^{12}$ equal to the size of the robot's degrees of freedom (DOF), and the torques applied to each joint for joint position control are calculated using PD control.

{\small
\bibliographystyle{unsrt}
\bibliography{egbib}
}

\addtolength{\textheight}{-12cm}   % 
                                  % 
%%%%%%%%%%%%%%%%%%%%%%%%%%%%%%%%%%%%%%%%%%%%%%%%%%%%%%%%%%%%%%%%%%%%%%%%%%%%%%%%

\end{document}